\documentclass{svproc}

\usepackage{doi}
\usepackage{hyperref}
\usepackage{amsmath}
\usepackage{booktabs}
\usepackage{multirow}
\usepackage{graphicx}

\usepackage{amssymb}
\usepackage{comment}

\begin{document}

\mainmatter

\title{Feature Engineered Reinforcement Learning for S-pair Selection in Buchberger's Algorithm}

\titlerunning{Feature Engineered RL for S-pair Selection in Buchberger's Algorithm}  

\author{Uzma Shafiq\inst{1,2} \and Matthew England\inst{1} \and Nayyar Zaidi\inst{2}
}

\authorrunning{Shafiq et al.} 

\tocauthor{Uzma Shafiq, Matthew England, Nayyar Zaidi}

\institute{Coventry University, UK \\ \email{shafiqu9@uni.coventry.ac.uk}, \email{matthew.england@coventry.ac.uk}, \and Deakin University, Australia  \\ \email{s224380786@deakin.edu.au, nayyar.zaidi@deakin.edu.au}
}

\maketitle              

\begin{abstract}
Selecting the order in which to process S-pairs is a crucial decision in Buchberger's algorithm for computing Gr\"obner bases: this choice can change the cost by orders of magnitude. Recent work has shown that reinforcement learning agents viewing exponent vectors of the polynomials outperform classical heuristics on binomial ideals. We extend this line of work to general polynomial ideals by using statistical features.  
We introduce a 17-dimensional feature vector to describe a polynomial and use PPO to train policies on five distributions of random ideals.  We find that using S-polynomial features significantly outperforms S-pair features on all five distributions. The new policies also outperform the retrained RL baseline on all five distributions, reducing algorithmic work by 41--53\% relative to the \textsc{degree} heuristic on four distributions where learning helps. The additional cost of the feature engineering means that wall-clock cost remains similar to the baseline, but our new approach benefits from easier transfer to unseen distributions.
\end{abstract}

\keywords{computer algebra, reinforcement learning, Buchberger's algorithm}

\section{Introduction}
\label{sec:Introduction}

Non-linear polynomial systems exist in a wide range of applications such as robotics \cite{AR_10}, cryptography \cite{FaugereJoux2003HFE}, 
physics \cite{MD_07}, algebraic geometry \cite{DK_86}, chemical reaction networks theory \cite{Sadeghimanesh_Feliu_2019} and more. Solving such systems exactly requires advanced computer algebra tools. One such tool is Gr\"{o}bner bases, which ``simplify'' systems of polynomial equations. Buchberger's algorithm  \cite{Buch_06} is the original procedure for computing Gr\"{o}bner bases and is still widely used. Despite decades of advancements, the runtime of Buchberger's algorithm in the worst case remains doubly exponential \cite{MEMA_1982}. It is also notoriously sensitive to the order in which to reduce a queue of pairs (so called S-pairs) of polynomials. Different selection strategies can yield runtimes that differ by orders of magnitude on the same input and there is no known optimal strategy for generic inputs. 
Classical selection heuristics, such as normal \cite{Buchberger_1979}, sugar \cite{Giovini_et_al_1991}, and degree \cite{PSH_20} predict which S-pairs are cheap to reduce.  These heuristics are fast to make decisions; however, they each use only one piece of information to make their decision.

Gr\"obner bases are used widely outside of pure algebra, and in most of those settings the difficulty is computational rather than conceptual. Algebraic cryptanalysis is one example. An attack on a cipher can often be written down as a polynomial system, such that the security of the system amounts to an argument that solving it is too expensive to attempt~\cite{FaugereJoux2003HFE}. Systems biology is another field of study where Gr\"obner bases are used widely. Under mass-action kinetics the steady states of a chemical reaction network are the solutions of a polynomial system: e.g., for the shuttle model of the Wnt signalling pathway this system has 19 unknowns and 36 parameters~\cite{Gross2016Wnt}. Models of that size are towards the limits of symbolic methods, and so analyses often fall back on numerical alternatives.

When symbolic methods fail they should not give a wrong answer; rather they run out of time, or more often memory. The latter happens because the polynomials produced during the process can become much larger than anything in the input or output (intermediate expression swell). How much they grow for Gr\"{o}bner bases depends on the order in which S-pairs are processed: so the choice of selection strategy often decides whether a problem can be solved at all.

There has been an increasing amount of work exploring the use of machine learning (ML) tools to optimise computer algebra algorithms. Topics include choosing the variable ordering for Cylindrical Algebraic Decomposition (CAD) (e.g. \cite{dRE24,PdREC24}), and  algorithm selection for symbolic integration (e.g. \cite{BEG24b,BSEG24}).
Peifer et al. \cite{PSH_20} were the first to use ML tools for optimising Gr\"{o}bner basis computation. In 2020, they used Reinforcement Learning (RL) to learn S-pair selection strategies. They trained a neural network to score pairs based on the exponent vectors of the polynomials. The work in that paper was limited to binomials (polynomials with only two terms), which is a crucial simplification since it firmly bounds the state space to binomials throughout the computation.  For more general polynomials, the intermediate calculations for Gr\"{o}bner basis calculation, and thus the state space representation size, can grow substantially.  
A preliminary experiment in the thesis \cite{Peifer2021} tested how the binomial-trained agent behaves on general polynomial ideals by truncating each polynomial to its first two monomials as a representation of the state at inference time. The agent beat the benchmarks on some ideals but behaved poorly on others. A full exploration of how to use RL to tackle this problem for general polynomials has not been addressed in the literature since.

This paper addresses that gap, showing that an RL methodology can be applied to general polynomials. However, we propose a feature engineered representation as the state rather than raw exponent vectors of the polynomials or their truncation. We build features from statistics derived from a polynomial. Crucially, because our representation is a set of statistics rather than raw polynomial data, we have a fixed-size feature vector regardless of how many terms a polynomial has: a polynomial with two terms, or fifty, or any other number is all encoded as a 17-dimensional feature set in our work. Thus, the models can be trained on wider distributions and applied more readily to distributions different from those on which they are trained.

We also examine the question of what the features should describe? Prior work (both the RL work in \cite{PSH_20} and the classical heuristics \cite{Buchberger_1979,Giovini_et_al_1991}) makes decisions based on the pair of polynomials to be processed. We will also explore making the decision using statistics about the S-polynomial itself (computed as part of that processing).  This will add to the cost of making the selection as some work has to be done up front, but it may provide more useful information to make the selection.  
We will compare these two feature representations as states in an identical RL framework, training Proximal Policy Optimization (PPO) \cite{schulman2017ppo} agents on both feature sets across five distributions of randomly generated general (non-binomial) polynomial ideals. We will evaluate against classical heuristics and an RL baseline based on \cite{Peifer2021}, reporting both the reward (total reduction cost) and wall-clock runtime for Gr\"{o}bner basis calculation.

We continue in Section \ref{sec:background} with background information on Buchberger's algorithm and then present our methodology and results in Sections \ref{sec:Methodology} and \ref{sec:results}, respectively.  We finish in Section \ref{sec:conclusion} with conclusions and ideas for future work.

\section{Background}
\label{sec:background}

\subsection{Buchberger's algorithm and pair selection}

We consider the polynomial ring $k[x_1,\ldots,x_n]$ where $k$ is our coefficient field and $x_i$ are variables.  An \emph{ideal} \( I \subseteq k[x_1,\ldots,x_n] \) is specified by a finite generating set of polynomials, $F=\{f_1,\ldots,f_s\}$, to represent the solutions to the system of equations formed by setting all $f_i=0$.  

A \emph{monomial order} on $k[x_1,\ldots,x_n]$ is a total well-ordering on the set of monomials from the ring that respects multiplication~\cite{Cox_et_al_2015}. The \emph{leading monomial} $\operatorname{LM}(f)$ and \emph{leading term} $\operatorname{LT}(f)$ of a polynomial $f$ are then its largest monomial with respect to the given ordering, and that monomial together with its coefficient, respectively. In this paper, we use the graded reverse lexicographic order throughout: order monomials by total degree first and then lexicographically.

The same ideal can have many different generating sets.  A generating set $G$ of an ideal $I$ such that the leading monomial (with respect to ordering) of every polynomial in $I$ is a multiple of the leading monomial of some polynomial in $G$ is called a \emph{Gr\"{o}bner basis}. This is a particularly useful generating set: it can simplify many computations and computationally extract properties of an ideal.  

Buchberger's algorithm computes a Gr\"{o}bner basis \( G \) of \( I \) by iteratively processing pairs of polynomials (called S-pairs). The set of pairs to process is initialised as all the pairs formed by the input generating set. For each pair \( (g_i,g_j) \) of basis elements, the algorithm forms the \(S\)-polynomial,
\[
S(g_i,g_j)
=
\frac{\operatorname{lcm}(\operatorname{LM}(g_i),\operatorname{LM}(g_j))}
{\operatorname{LT}(g_i)}
g_i
-
\frac{\operatorname{lcm}(\operatorname{LM}(g_i),\operatorname{LM}(g_j))}
{\operatorname{LT}(g_j)}
g_j,
\]
where lcm is the least common multiple of the two monomials.  The algorithm then reduces \( S(g_i,g_j) \) modulo the current basis \( G \). If the reduction is non-zero, the remainder is added to \( G \) and new pairs are formed from the remainder and the rest of $G$. The process terminates when all pending pairs have been processed. To get a detailed understanding of the algorithm please refer to \cite{Buch_06}.

The order in which pairs are processed is critical in determining both the size of intermediate computations and the time required for the algorithm to terminate.  The S-pair selection problem is the choice, at each step, of which pair from the pending queue to process next. Classical strategies include:

\begin{itemize}
    \item \textbf{First:} process pairs simply in the order in which they were generated.

    \item \textbf{Normal} \cite{Buchberger_1979}: process the pair whose \(\operatorname{lcm}\) of leading monomials is smallest with respect to the monomial order.

    \item \textbf{Degree} \cite{PSH_20}: process the pair with the smallest total degree of that \(\operatorname{lcm}\).

    \item \textbf{Sugar} \cite{Giovini_et_al_1991}: process the pair with the smallest ``sugar'', a virtual degree that tracks what the \(\operatorname{lcm}\)-degree would be if the inputs had been homogeneous.
\end{itemize}

These heuristics seek to predict the size of the S-polynomial, and are deterministic, computationally inexpensive to evaluate, and theoretically motivated.  However, they do not always make good decisions.

\subsection{Learned pair selection}

Peifer et al. \cite{PSH_20} model pair selection as a Markov decision process in which the state is the exponents of the two monomials of the two binomials in the current pair queue, the action is the choice of pair to reduce next, and the reward is negative, i.e. a cost: the negative of the number of polynomial additions of the resulting reduction. For a pair \( (g_i,g_j) \) in a polynomial ring in \(n\) variables, the state is derived from the \(n\)-dimensional exponent vectors of both binomials in the pair. When calculating Gr\"{o}bner bases for a binomial ideal we are guaranteed that any intermediate polynomials calculated are also binomials. This means our state is represented by fixed-shape tensors, the dimensions of which depends only on \(n\), not on the binomials themselves.  This fixed size integrates cleanly with neural architectures that expect fixed-shape inputs. 
Peifer et al. \cite{PSH_20} trained a policy network using PPO \cite{schulman2017ppo} on a distribution of randomly generated binomial ideals and showed that the learned policy generalises to held-out ideals from the same distribution and outperforms classical selection heuristics.

This state representation has a structural limitation.  While the reduction of an \(S\)-polynomial of two binomials produces another binomial, there is no such result for more general polynomials.  \(S\)-polynomials of general polynomials can have significantly  many more terms than the polynomials in the original pair, resulting in an unpredictable evolution of the size and structure of intermediate polynomials as the algorithm runs. The published work \cite{PSH_20} did not address this limitation, but the thesis \cite{Peifer2021} noted that the agent could be applied to more general polynomials simply by truncating each polynomial to its first two monomials as a representation of the state. Preliminary experiments in the thesis suggested that the agent beat the benchmarks on some ideals but behaved poorly on others. This approach is clearly limited: a polynomial with two terms and a polynomial with the same two first in the ordering but then many more can result in very different reduction costs. 

To provide a fair RL baseline for general polynomial ideals, we retrain Peifer's framework on our five general polynomial distributions with $k=3$ monomials per polynomial: the smallest extension beyond Peifer's published binomial distributions, which also matches the typical term count of ideals of our generator. Thus, what we refer to as Peifer's RL baseline throughout the paper truncates the polynomials like this to have a fixed dimension for the state representation.

\subsection{Hypothesis}

Our work follows the same RL framework as \cite{PSH_20}, using PPO over a Markov Decision Process (MDP) for pair selection, but replaces the state representation with distributional statistics computed over the full polynomial, thus making the resulting state vector fixed-size regardless of how many terms a polynomial contains. 
We investigate two hypotheses.
\begin{description}
    \item[\textbf{H1:}] Learned policies using \(S\)-polynomial-based features outperform learned policies using \(S\)-pair-based features on the pair-selection task. The \(S\)-polynomial is the object the algorithm reduces; statistics computed from it should be more predictive of reduction cost than statistics computed from its inputs.

    \item[\textbf{H2:}] A learned policy using distributional polynomial statistics outperforms both (i) classical pair-selection heuristics (first, degree, normal, sugar) and (ii) the leading-monomial-based learned policy of Peifer et al. \cite{PSH_20}, on randomly generated general polynomial ideals.
\end{description}

We will test both hypotheses on five different distributions to observe how results change with each parameter of our random ideal generator, measuring both reduction cost (the reward signal used during training) and wall-clock runtime to compute the corresponding Gr\"{o}bner basis.

\section{Methodology}
\label{sec:Methodology}

We use the RL framework for pair selection introduced by Peifer et al. \cite{PSH_20}: pair selection is cast as a MDP, and a policy is trained using PPO \cite{schulman2017ppo} to minimise total reduction cost (see \cite{PSH_20} for the full MDP formulation).  
The reward at each step is $-(1+k)$, where $k$ is the number of polynomial-addition operations performed during the reduction of the S-polynomial of the chosen pair; so cumulative episode reward equals the negative of the total reduction work. Episodes terminate when the pair queue is empty; the Gr\"{o}bner basis thus computed. 

\subsection{Dataset}
We synthetically generate five different dataset distributions following the data generation process used in \cite{PSH_20}. Each dataset consists of randomly generated polynomials defined over a fixed coefficient field $\mathbb{Z}/32003\mathbb{Z}$. The datasets have four different parameters denoted \((v,\,d,\,p,\,\sigma)\), where \(v\) is the number of variables, \(d\) is the maximum total degree, \(p\) is the number of polynomials per system, and  \(\sigma\) is a density parameter used to determine the number of terms generated for each polynomial. In this paper, we keep $\sigma$ constant at 0.3 (holding the density per-polynomial constant) but vary the algebraic-shape parameters $(v, d, p)$ and thus use these three numbers to specify the five datasets later.

\subsection{Feature representation}

The state visible to the policy is a feature matrix of shape $P \times d$, where $P$ is the number of pairs to process and $d$ is the number of features.  Our central design choice is a fixed vocabulary of $17$ distributional statistics applicable to any polynomial, as summarised in Table \ref{tab:poly_statistics}. 
The size of this set of features is invariant with the shape of the actual polynomials. A polynomial with two terms or more will have the same state size regardless of the number of monomials. This representation allows us to apply the approach to general polynomials. We instantiate the features in two different manners:
\begin{itemize}
\item \textbf{S-polynomial based} ($d=17$): for each pair \( (g_i,g_j) \) in P we compute the S-polynomial \(S( g_i,g_j) \) and then compute the 17 statistics in Table \ref{tab:poly_statistics}. 

\item \textbf{S-pair based} ($d=34$): For each pair  \( (g_i,g_j) \) we compute the 17 statistics for both  \( g_i \) and  \( g_j \) and concatenate them. 
\end{itemize}

\begin{table}[t]
\centering
\caption{Polynomial statistics used as features}
\label{tab:poly_statistics}
\begin{tabular}{c l l}
\hline
\textbf{\#} & \textbf{Statistic} & \textbf{Definition} \\
\hline

1 & degree\_max & maximum total degree across all terms in the polynomial \\
2 & degree\_min & minimum total degree \\
3 & degree\_avg & mean total degree \\
4 & degree\_std & standard deviation of total degrees \\
5 & degree\_range & \( \text{degree\_max} - \text{degree\_min} \) \\
6 & degree\_median & median total degree \\
7 & num\_terms & number of nonzero terms \\
8 & degree\_entropy & Shannon entropy of degree distribution \\
9 & lm\_degree & total degree of the leading monomial \\
10 & tail\_degree & total degree of the smallest monomial under the order \\
11 & lm\_tail\_gap & \( \text{lm\_degree} - \text{tail\_degree} \) \\
12 & density & 
$\text{num\_terms} / \binom{n+\text{degree\_max}}{n}$
\\
13 & coeff\_entropy & Shannon entropy over normalised coefficient magnitudes \\
14 & coeff\_max & maximum absolute coefficient \\
15 & coeff\_std & standard deviation of absolute coefficients \\
16 & degree\_skew & sample skewness of degree distribution \\
17 & degree\_kurtosis \qquad & sample excess kurtosis of degree distribution\\
\hline
\end{tabular}
\end{table}

\subsection{Policy and training}

For a variable-size action space, the queue of S-pairs grows and shrinks across the episodes. To handle this, the policy applies a shared multi-layer perceptron (MLP) independently to each row of the feature matrix, and yields a scalar score per pair, followed by a softmax over these scores. The MLP has architecture $d \rightarrow 128 \rightarrow 64 \rightarrow 1$ with ReLU activations.  The value network has an identical architecture with mean-pooling over pairs before the final scalar output. 
We train both networks with PPO \cite{schulman2017ppo}, using discount factor $\gamma = 0.99$, GAE parameter $\lambda = 0.97$, clip ratio $\epsilon = 0.2$, and KL early-stopping at $0.01$. Optimisation uses Adam with policy and value learning rates $10^{-4}$ and $10^{-3}$ respectively. 
Each training epoch consists of rollouts on $250$ randomly generated ideals, followed by $10$ policy updates and $25$ value updates. The ideals are generated using a random seed of $42$. Training continues until convergence.

\subsection{Evaluation protocol}

For each training distribution, we generate a test set of 2000 ideals using a distinct seed. Every selection strategy, classical heuristics and learned policies, are evaluated on the same test sets. For each pair (strategy, ideal), we record  the total reward, episode length, and wall-clock runtime including feature extraction.

\section{Results}
\label{sec:results}

We evaluate the classical and learned selection strategies on five  randomly generated distributions of general polynomial ideals. The four classical heuristics that we used were first, degree, normal, sugar; while the learned strategies evaluated were the trained policy from Peifer et al. \cite{PSH_20} \cite{Peifer2021}, and our two distributional-feature variants (S-pair and S-poly). Each strategy was tested on the same 2000 held-out test ideals per distribution, generated from a fixed seed distinct from training. 
We present results to answer the two research questions separately.

\subsection{H1: S-polynomial vs S-pair feature representation}
\label{sec:RQ1}

Table \ref{tab:rq1} reports the comparison between our two different feature designs (S-polynomial and S-pair based) on identical training and evaluation pipelines. 
The S-poly representation outperforms on all five distributions, despite using half the number of features. The improvement is statistically significant (Wilcoxon $p < 0.01$) on all five distributions.  Relative improvements range from $3.2\%$ on 3-15-10 to $28.9\%$ on 3-10-3 and $21.9\%$ on 4-5-10. S-poly is never worse than the S-pair in mean reward and is significantly better in every distribution we tested.

\begin{table}[h]
\centering
\caption{S-poly vs S-pair feature representations. Mean, median, and
standard deviation of reward across 2{,}000 held-out ideals. Wilcoxon $p$ is paired. Bold marks the better representation per
metric.}
\label{tab:rq1}
\small
\setlength{\tabcolsep}{4pt}
\begin{tabular}{lrrrrrrc}
\toprule
& \multicolumn{2}{c}{Mean $R$} & \multicolumn{2}{c}{Median $R$} & \multicolumn{2}{c}{Std $R$} & \\
\cmidrule(lr){2-3}\cmidrule(lr){4-5}\cmidrule(lr){6-7}
Dist. & S-pair & S-poly & S-pair & S-poly & S-pair & S-poly & $p$ \\
\midrule
3-5-10  & $-59.4$  & $\mathbf{-48.5}$  & $-52.0$  & $\mathbf{-42.0}$  & $40.3$   & $\mathbf{30.6}$   & $< 10^{-100}$ \\
3-10-3  & $-308.7$ & $\mathbf{-219.5}$ & $\mathbf{-56.0}$ & $-59.0$ & $1428.7$ & $\mathbf{891.3}$  & $4.6\times 10^{-12}$ \\
3-10-10 & $-173.0$ & $\mathbf{-148.0}$ & $-93.0$  & $\mathbf{-88.0}$  & $421.6$  & $\mathbf{332.8}$  & $2.3\times 10^{-15}$ \\
3-15-10 & $-663.4$ & $\mathbf{-642.1}$ & $-155.0$ & $\mathbf{-146.0}$ & $\mathbf{2473.1}$ & $2960.7$ & $9.5\times 10^{-8}$ \\
4-5-10  & $-136.6$ & $\mathbf{-106.7}$ & $-93.0$  & $\mathbf{-68.0}$  & $209.6$  & $\mathbf{203.2}$  & $< 10^{-100}$ \\
\bottomrule
\end{tabular}
\end{table}

Another observation is that S-poly yields lower per-ideal variance than S-pair on four of the five distributions. The only exception is distribution 3-15-10, where S-poly's standard deviation is marginally higher despite a better mean. The two policies were trained with the same architecture and training hyperparameters; therefore  this difference reflects the value of the feature representation itself. Features computed directly from $S(f,g)$ give information about the object that the algorithm will reduce, while features computed from $f$ and $g$ separately give indirect information to predict reduction cost.

These results support hypothesis H1: the S-poly representation is at least as effective as the S-pair representation across all five tested distributions in terms of the reward it was trained for, with no instance of S-pair winning.

\subsection{H2: Learned policies vs Peifer and classical heuristics}
\label{sec:rq2}

We compare the S-poly policy, our learned method based on Section \ref{sec:RQ1} against the policy trained using an exponent vector of $k=3$ monomials following \cite{Peifer2021} and four classical heuristics.
Table~\ref{tab:rq2_reward} reports reward statistics across all five distributions, while Table~\ref{tab:rq2_time} reports the corresponding total wall-clock cost.

\begin{table}[b]
\centering
\caption{Mean reward $\pm$ standard deviation per strategy across five distributions of 2{,}000 held-out ideals. Higher (less negative) reward is better; best per row in bold.}
\label{tab:rq2_reward}
\setlength{\tabcolsep}{3.5pt}
\resizebox{\textwidth}{!}{%
\begin{tabular}{lrrrrrrrr}
\toprule
Distribution & \textsc{random} & \textsc{first} & \textsc{degree} & \textsc{normal} & \textsc{sugar} & \textsc{peifer} & \textsc{S-pair} & \textsc{S-poly} \\
\midrule
3-5-10  & $-98 \pm 106$    & $-96 \pm 98$    & $-64 \pm 58$    & $-61 \pm 49$         & $-72 \pm 70$    & $-58 \pm 44$    & $-59 \pm 40$    & $\mathbf{-49 \pm 31}$  \\
3-10-3  & $-624 \pm 3046$  & $-331 \pm 1941$ & $-197 \pm 804$  & $\mathbf{-196 \pm 632}$ & $-209 \pm 740$ & $-233 \pm 990$ & $-309 \pm 1429$ & $-220 \pm 891$ \\
3-10-10 & $-1034 \pm 2024$ & $-732 \pm 1400$ & $-312 \pm 824$  & $-278 \pm 708$       & $-444 \pm 905$  & $-172 \pm 434$  & $-173 \pm 422$  & $\mathbf{-148 \pm 333}$ \\
3-15-10 & $-6421 \pm 14118$ & $-3646 \pm 7873$ & $-1322 \pm 4872$ & $-1176 \pm 4676$    & $-1930 \pm 4661$ & $-795 \pm 3676$ & $-663 \pm 2473$ & $\mathbf{-642 \pm 2961}$ \\
4-5-10  & $-516 \pm 1325$  & $-336 \pm 609$  & $-182 \pm 382$  & $-157 \pm 271$       & $-215 \pm 399$  & $-134 \pm 242$  & $-137 \pm 210$  & $\mathbf{-107 \pm 203}$  \\
\bottomrule
\end{tabular}%
}
\end{table}

\paragraph{Learned methods vs classical heuristics.}
On four of the five distributions, 3-5-10, 3-10-10, 3-15-10, and 4-5-10, all learned methods (Peifer, S-pair and S-poly) outperform every classical heuristic. The improvement grows monotonically as the maximum polynomial degree $d$ increases. The S-poly policy improves over degree by $+24.5\%$ on 3-5-10, $+52.6\%$ on 3-10-10, and $+51.4\%$ on 3-15-10. We can also observe a significant difference in the standard deviation on 3-15-10, which is 54872 for degree and 4676 for normal, compared with 2961 for S-poly (less than 61\% of degree's spread). This shows that the learned policies not only improve average performance, but also avoid failure cases that result in inflation of the heuristic's variance on harder problems.

The only exception to the dominance of the learned strategies on this metric is with distribution 3-10-3,
where all three learned methods underperform normal selection 
This is not a failure of just one learned policy, but every learned method we tested. With the $s=3$ initial generators, the pairs to choose from the queue are only 3--5 pairs at any decision step. This leaves the action space constrained for a learned policy to extract meaningful benefit over classical heuristics. 

\paragraph{S-poly vs Peifer's strategy.}

Peifer's exponent vector representation was originally developed and tested for binomial ideals in \cite{PSH_20}. Then in \cite{Peifer2021} it was suggested to apply this more generally by representing each polynomial by the exponent vector of its $k$ leading monomials with respect to the ordering, padding with zeros if it has fewer terms. The pair's feature is the concatenation of these vectors for $f$ and $g$, giving a $2nk$-dimensional input. We evaluated using $k=3$ for Peifer's policies, a small extension of Peifer's preliminary experiments with $k=2$.  Despite discarding much information, it works surprisingly well in Table \ref{tab:rq2_reward}. Peifer's policy beats every classical heuristic on the four distributions (e.g., $p = 5.2 \times 10^{-63}$ vs degree on 3-10-10), placing it second or third (depending on the distribution) among the strategies tested.

The S-poly outperforms Peifer's methodology in mean reward on four of five distributions, which is statistically significant on four of them: $+15.9\%$ on 3-5-10 ($p = 3.6 \times 10^{-94}$), $+5.9\%$ on 3-10-3 ($p = 0.040$), $+19.2\%$ on 3-15-10 ($p = 1.4 \times 10^{-19}$), and $+20.2\%$ on 4-5-10 ($p = 7.8 \times 10^{-84}$). On 3-10-10 the mean gap is $+14.0\%$ ($-148.0$ vs $-172.2$) but not statistically significant  ($p = 0.115$). 
The results across distributions suggest that the advantage of proposed features over leading-monomial exponent vectors is consistent but variable in magnitude.

\paragraph{Wall-clock cost.}
The wall-clock cost for the different policies is given in Table \ref{tab:rq2_time}.
Classical heuristics are the cheapest in terms of wall-clock runtime: leading to normal selection outperforming on two of five distributions (3-5-10 and 3-10-3). On two of the distributions, Peifer's learned policy outperforms other strategies (3-10-10 and 4-5-10). Among learned methods, Peifer's method is the cheapest, since the state representation just reads exponent vectors that the polynomial representation already caches. On 3-15-10, S-pair is the fastest learned method.

\begin{table}[b]
\centering
\caption{Total wall-clock time (seconds) to solve 2{,}000 held-out ideals per
distribution, including feature extraction and policy
inference; fastest on each row is in bold.}
\label{tab:rq2_time}
\setlength{\tabcolsep}{4pt}
\small
\begin{tabular}{lrrrrrrrr}
\toprule
Dist. & \textsc{random} & \textsc{first} & \textsc{degree} & \textsc{normal} & \textsc{sugar} & \textsc{peifer} & \textsc{S-pair} & \textsc{S-poly} \\
\midrule
3-5-10  & $2.83$ & $2.63$ & $2.12$ & $\mathbf{2.09}$ & $2.39$ & $2.53$ & $3.36$ & $3.09$ \\
3-10-3  & $109.22$ & $43.43$ & $21.58$ & $\mathbf{18.74}$ & $20.82$ & $28.00$ & $38.74$ & $28.58$ \\
3-10-10 & $52.26$ & $31.66$ & $14.32$ & $12.56$ & $18.91$ & $\mathbf{8.27}$ & $10.26$ & $10.05$ \\
3-15-10 & $722.60$ & $328.65$ & $108.01$ & $97.05$ & $147.55$ & $62.23$ & $\mathbf{58.25}$ & $61.18$ \\
4-5-10  & $21.75$ & $10.64$ & $6.77$ & $6.09$ & $7.97$ & $\mathbf{6.04}$ & $7.98$ & $7.16$ \\
\bottomrule
\end{tabular}
\end{table}

S-poly and S-pair are competitive with Peifer and end up with a small margin of difference in wall-clock cost. The S-poly to Peifer time ratio is $1.22\times$ on 3-5-10, $1.02\times$ on 3-10-3, $1.21\times$ on 3-10-10, $0.98\times$ on 3-15-10, and $1.19\times$ on 4-5-10. 
On 3-15-10 the S-pair is narrowly ahead of Peifer ($62.2$\,s) and the S-poly ($61.2$\,s). Computing $S(f,g)$ before extracting its 17 statistics adds overhead, but this is outweighed on harder distributions by better pair choices.
 
Together, Tables~\ref{tab:rq2_reward} and \ref{tab:rq2_time} shows a small trade-off between reward and wall clock time. On 3-10-10 Peifer is at $(8.27\,\text{s},\,-172.2)$, the S-pair is at $(10.26\,\text{s}$, $-173.0)$, and the S-poly is at $(10.05\,\text{s},\,-148.0)$. The S-poly delivers $14\%$ better mean reward than Peifer at $1.21\times$ the total time; on 3-15-10 the corresponding numbers are $19\%$ better reward at $0.98\times$ the time, i.e. better in both. 

These results support hypothesis H2 on reward but not on wall-clock cost:
S-poly outperforms both classical heuristics and Peifer in mean reward on four of five distributions, but is $1.02-1.25\times$ slower than Peifer in total wall-clock time. The reward improvement comes at a small but persistent overhead cost. 

\subsection{Cross-distribution transfer}
\label{sec:transfer}

We further tested if the encoded features are generalisable for selecting out of the distribution on which they were trained. We evaluate the S-poly and S-pair policies trained on the 3-10-10 distribution on the other four test distributions (3-5-10, 3-10-3, 3-15-10, 4-5-10), without any fine-tuning. Note that distribution 4-5-10 involves $n=4$ variables, one more than the training distribution; this transfer in the case of Peifer would not be possible due to the architectural limitation (dimensionality is tied to $n$) however, it is possible in our proposed feature engineered approach because the 17 polynomial statistics are dimension-agnostic (do not depend on the variable count).
Table~\ref{tab:transfer} reports the mean rewards of transferred policies alongside two in distribution reference points for each test distribution: the Peifer policy trained on that distribution, and the better of the two strongest classical heuristics (normal, degree).

\begin{table}[b]
\centering
\caption{To evaluate cross-distribution transfer S-poly and S-pair policies trained
on 3-10-10 are evaluated on four unseen distributions. Mean reward
$\pm$ standard deviation over 2{,}000 held-out ideals per distribution. Transferred policies are denoted with subscript
$T$ and in distribution policies are denoted with subscript id. Peifer and the best heuristic on each distribution are reproduced
from Table~\ref{tab:rq2_reward} for comparison. The best mean in each row is in bold.}
\label{tab:transfer}
\small
\setlength{\tabcolsep}{5pt}
\begin{tabular}{lrrrr}
\toprule
Test dist. & S-poly$_T$ & S-pair$_T$ & Peifer$_\mathrm{id}$ & Best heuristic \\
\midrule
3-5-10  & $-66.2 \pm 60$            & $-67 \pm 52$    & $\mathbf{-57 \pm 43}$ & $-61 \pm 48$ \\
3-10-3  & $\mathbf{-188 \pm 745}$ & $-257 \pm 1182$ & $-233 \pm 990$       & $-196 \pm 631$ \\
3-15-10 & $\mathbf{-402 \pm 1820}$ & $-591 \pm 2306$ & $-794 \pm 3676$     & $-1175 \pm 4675$ \\
4-5-10  & $-174 \pm 363$          & $-175 \pm 291$  & $\mathbf{-133 \pm 241}$ & $-157 \pm 271$ \\
\bottomrule
\end{tabular}
\end{table}

The transfer results reveal three findings, each with implications for the practical use of learned pair selection.

\emph{Transfer succeeds to a higher degree.}
On 3-15-10 and 3-10-3, the transferred S-poly policy not only performs well but it also  outperforms the corresponding in-distribution Peifer baseline: by $19.4\%$ on 3-10-3 ($-188.0$ vs $-233.2$, $p = 3.9 \times 10^{-43}$) and by $49.3\%$ on 3-15-10 ($-402.7$ vs $-794.9$, $p = 2.3 \times 10^{-41}$). A policy that has never observed a 3-15-10 ideal still beats a Peifer policy trained directly on that distribution! The transferred policy also substantially beats the best classical heuristic on 3-15-10 (normal, $+65.7\%$ improvement, $p < 10^{-100}$) and marginally beats it on 3-10-3 (normal, $+4.1\%$ improvement, $p = 6.1 \times 10^{-8}$).

\emph{Transfer degrades on lower degree and higher variables.}
On the distributions 3-5-10 and 4-5-10, the transferred policy underperforms the corresponding Peifer's policies and falls slightly behind the normal heuristic in mean reward. The smaller degree polynomials in these distributions may yield features the policy did not see during training so decision boundaries do not transfer well.

\emph{S-poly retains its advantage over S-pair under transfer.}
The relative ranking between S-poly and S-pair features is preserved on all four test distributions: transferred S-poly significantly outperforms transferred S-pair on every one. This confirms that the S-polynomial features' advantage over the S-pair features found in Section~\ref{sec:RQ1} is maintained under transfer.

\section{Conclusion and Future Work}
\label{sec:conclusion}

We investigated two questions about feature design for learned pair selection in Buchberger's algorithm. For \textbf{H1} (S-polynomial vs S-pair features), the S-polynomial representation significantly outperformed the S-pair variant on all the distributions, supporting the hypothesis that features computed from the object the algorithm will actually reduce are more informative. For \textbf{H2} (learned policies vs classical heuristics and Peifer's leading-monomial baseline), our S-polynomial policy outperformed every classical heuristic on four of the distributions, with mean cost reductions of up to 53\%, and outperformed Peifer's leading-monomial features in mean reward on all five distributions (significantly on four). However, in terms of wall-clock cost, S-poly policy is $1.02$ to $1.25\times$ slower than Peifer, i.e. the overhead of the feature engineered approach outweighs its benefits for these distributions.  However, the value of the new approach comes also from its ease and ability to generalise to out-of-distribution datasets. 

For future work, there are two natural directions: extending the evaluation to higher variable counts, degrees, and real benchmark ideals to test whether the observations persist at scale. It would also be interesting to study how polynomial density affects
these policies and a hybrid policy that selects between cheap leading-monomial features and the more expensive S-polynomial features per pair, to recover the reward advantage at lower wall-clock cost. 


\subsubsection{\ackname} US is supported by a joint scholarship from Coventry University and Deakin University.

\bibliographystyle{splncs04}
\bibliography{references}

\begin{thebibliography}{6}
%

\bibitem {smit:wat}
Smith, T.F., Waterman, M.S.: Identification of common molecular subsequences.
J. Mol. Biol. 147, 195?197 (1981). \url{doi:10.1016/0022-2836(81)90087-5}

\bibitem {may:ehr:stein}
May, P., Ehrlich, H.-C., Steinke, T.: ZIB structure prediction pipeline:
composing a complex biological workflow through web services.
In: Nagel, W.E., Walter, W.V., Lehner, W. (eds.) Euro-Par 2006.
LNCS, vol. 4128, pp. 1148?1158. Springer, Heidelberg (2006).
\url{doi:10.1007/11823285_121}

\bibitem {fost:kes}
Foster, I., Kesselman, C.: The Grid: Blueprint for a New Computing Infrastructure.
Morgan Kaufmann, San Francisco (1999)

\bibitem {czaj:fitz}
Czajkowski, K., Fitzgerald, S., Foster, I., Kesselman, C.: Grid information services
for distributed resource sharing. In: 10th IEEE International Symposium
on High Performance Distributed Computing, pp. 181?184. IEEE Press, New York (2001).
\url{doi: 10.1109/HPDC.2001.945188}

\bibitem {fo:kes:nic:tue}
Foster, I., Kesselman, C., Nick, J., Tuecke, S.: The physiology of the grid: an open grid services architecture for distributed systems integration. Technical report, Global Grid
Forum (2002)

\bibitem {onlyurl}
National Center for Biotechnology Information. \url{http://www.ncbi.nlm.nih.gov}
\end{thebibliography}

%

\end{document}